# Patch-Wise Self-Supervised Visual Representation Learning: A Fine-Grained Approach

Ali Javidani[1*], Mohammad Amin Sadeghi[2], Babak Nadjar Araabi[1]

[1] School of Electrical and Computer Engineering, College of Engineering, University of Tehran, Tehran, Iran

[2] Qatar Computing Research Institute, Doha, Qatar

*alijavidani@ut.ac.ir

*Abstract-* Self-supervised visual representation learning traditionally focuses on image-level instance discrimination. Our study introduces an innovative, fine-grained dimension by integrating patch-level discrimination into these methodologies. This integration allows for the simultaneous analysis of local and global visual features, thereby enriching the quality of the learned representations. Initially, the original images undergo spatial augmentation. Subsequently, we employ a distinctive photometric patch-level augmentation, where each patch is individually augmented, independent from other patches within the same view. This approach generates a diverse training dataset with distinct color variations in each segment. The augmented images are then processed through a self-distillation learning framework, utilizing the Vision Transformer (ViT) as its backbone. The proposed method minimizes the representation distances across both image and patch levels to capture details from macro to micro perspectives. To this end, we present a simple yet effective patch-matching algorithm to find the corresponding patches across the augmented views. Thanks to the efficient structure of the patch-matching algorithm, our method reduces computational complexity compared to similar approaches. Consequently, we achieve an advanced understanding of the model without adding significant computational requirements. We have extensively pretrained our method on datasets of varied scales, such as Cifar10, ImageNet-100, and ImageNet-1K. It demonstrates superior performance over state-of-the-art self-supervised representation learning methods in image classification and downstream tasks, such as copy detection and image retrieval. The implementation of our method is accessible on GitHub.

*Keywords*- Self-Supervised Learning; Patch-Wise Representation Learning; Self-Distillation; Patch-level Augmentation; Patch-Matching.

## 1. Introduction

The concept of instance discrimination, integral to the evolution of self-supervised representation learning, treats each sample in a mini-batch as a unique class, thereby distancing their representations from one another [1-10]. This approach has been widely recognized and applied in prominent frameworks such as SimCLR [9] and MoCo [8]. Despite its popularity, instance discrimination faces several challenges: the presumption of distinct classes for each mini-batch sample may not hold true, given the possibility of samples sharing the same concept or label; the method's $O(n^2)$ time complexity complicates the training process; and it often fails to capture low-level features critical for detailing an image.

To address these drawbacks, various strategies have been developed. Clustering-based methods like SwAV [11], PCL [12], and oBoW [13], which incorporate online clustering into representation learning, aim to mitigate the issue of misclassifying conceptually similar samples, thereby enhancing contrastive method performance [11-15]. Non-contrastive approaches such as BYOL [16], SimSiam [17], and DINO [18] avoid the pitfalls of representation divergence without compromising training efficiency [16-24]. However, the challenge of capturing low-level features remains largely unaddressed, with most research focusing on image-level characteristics. This oversight leads to a potential neglect of critical details indicating an ongoing need for research in this area.

In this paper, we aim to enhance the granularity with which models understand image details through self-supervised learning, introducing a novel approach that extends consistency loss application beyond augmented images to include their internal local regions, such as patches. This extension produces additional loss terms, aiding in a deeper comprehension of image intricacies. Our key contributions are:

- Patch-level augmentation: We introduce the novel concept of augmenting patches independently of each other and the entire image, a method previously unexplored and offering new avenues for research in self-supervised learning.
- Patch-matching algorithm: a simple yet effective patch-matching algorithm that finds the corresponding patches across the augmented views in $O(1)$. It can be highly useful in self-supervised representation learning frameworks.

- PW-Self: a patch-wise self-supervised visual representation learning model that uniquely integrates local and global loss calculations into several cohesive loss functions, enabling a more comprehensive understanding of visual data at both micro and macro levels.
- Compare the proposed PW-Self method with other approaches, achieving state-of-the-art results on small, medium, and large-scale datasets.

PW-Self distinguishes itself from existing patch-wise self-supervised representation learning methods by addressing key limitations observed in the literature [25-27]. Unlike SelfPatch, which applies an invariance constraint to a patch and its neighbors without considering the possibility of conceptual divergence among adjacent patches or their inability to differentiate distinct object parts [26], PW-Self enhances learning by aligning representations of the same object parts across augmented images. EsViT proposes a multi-stage architecture utilizing sparse self-attention and a region-matching pretraining task based on cosine similarity to capture fine-grained features [27]. However, this method faces challenges in reliably identifying the most relevant patch during the network's early learning phase and incurs a high computational cost with its $O(T^2)$ time complexity (where $T$ is the number of patches). PW-Self overcomes these issues with an efficient patch-matching algorithm that operates in $O(1)$ time, ensuring the accurate pairing of corresponding patches. Furthermore, while some approaches push away representations of local views from the same image under the assumption of dissimilar content [25], PW-Self instead focuses on recognizing and bringing closer the representations of similar local regions across different views, thereby offering a more nuanced understanding of image content.

The structure of this paper is outlined as follows: Section 2 provides a review of literature relevant to our approach. Section 3 offers an in-depth explanation of the proposed PW-Self method and its underlying theory. In Section 4, we present the results achieved by PW-Self, including a comparative analysis with leading methods across different datasets. Section 5 delves into discussions about the computational complexities and the impact of parameters pertinent to our method.

## 2. Related Works: Self-Supervised Representation Learning

The evolution and enhancement of contrastive methods in machine learning have been marked by significant developments. Initially, SimCLR emerged as a pioneering approach utilizing contrastive loss, focusing on instance classification by contrasting augmented images from the same source against those from different sources [9]. It evolved into SimCLR-v2 [28], which incorporated extensive unlabeled data for pretraining and labeled data for fine-tuning, although its effectiveness is constrained by mini-batch size, posing memory challenges. MoCo [8] offered a novel approach with separate query and key branches, leading to its iterations MoCo-v2 [29] and MoCo-v3 [30], the latter integrating ViT for stability. Contrastive Predictive Coding (CPC) [10] and Self-Distilled Self-Supervised Learning (SDSSL) [31] further diversified the field, with CPC predicting future data in latent space and SDSSL using self-distillation for learning intermediate representations. More recent methods have shifted focus towards optimizing contrastive learning by challenging negative sample selection. Strategies like hard negative mixing at the feature level [7], reduction of mutual information between views [32], and object localization in ContrastiveCrop aim to improve contrastive views [33]. NNCLR enhances performance by redefining nearest samples as positives [4], while other strategies focus on difficult negative sample selection [1, 2] and hard negative example generation [34]. Addressing the issue of false negatives in mini-batches, some methods propose cancellation and attraction strategies [35]. TMCT introduces a third network with momentum for target learning [36].

The contrastive approach, despite its advancements, encounters challenges due to instance discrimination. The issue is particularly problematic when images of the same class are in the same mini-batch, leading to undesirable representation divergence. To address this, clustering has been integrated into contrastive learning frameworks [11-15]. The initial steps in this direction include Deep Embedded Clustering (DEC) [37], which combines an autoencoder with clustering in latent space, and Caron *et al.*'s method of using a Convolutional Neural Network for feature extraction and clustering to provide pseudo-labels for network training [15]. Recent advancements have furthered this integration. Techniques like Sinkhorn-Knopp transform clustering after feature extraction [11], Prototypical Contrastive Learning (PCL) that employs prototypes as centroids for image clusters [12], the OBoW method using a teacher-student network for visual word prediction [13], and the MCVT framework that merges cluster-based and info-based contrastive losses [38], all contribute to overcoming the limitations of contrastive learning. Additionally, the Self-Supervised Pyramid Representation Learning (SS-PRL) offers pyramid representations at the patch level, employing supplementary learners for semantic analysis [39]. These innovations illustrate a concerted effort to refine contrastive learning by incorporating clustering techniques, thereby enhancing the performance and accuracy of self-supervised representation learning systems.

Non-contrastive learning methods, focusing exclusively on positive samples, diverge from contrastive approaches by not requiring negative pair comparisons. BYOL (Bootstrap Your Own Latent) exemplifies this by utilizing two views from a single image, processed through encoders to minimize the mean squared error between the online and a target encoder, the latter being an exponential moving average of the former [16]. An advancement over BYOL, MYOW, employs two distinct but closely related samples, achieving superior results by avoiding repetitive augmentations [23]. DINO enhances this paradigm through self-distillation with a teacher-student network, both utilizing the Vision Transformer architecture and incorporating momentum, centering, and sharpening modules for the teacher network to avoid trivial outcomes [18]. MSBReg builds upon BYOL by integrating three distinct loss terms that harness multiple data sources for enhanced performance [40]. SimSiam [17] further streamlines the approach by eschewing negative samples, clustering, and moving averages, offering computational efficiency and superior performance in scenarios with smaller batch sizes, outpacing counterparts like SimCLR [9] and BYOL [16].

Recent developments in non-contrastive learning focus on maximizing the information within representation vectors to prevent representation collapse, a challenge noted in earlier research [41]. These methods employ specific constraints in their objective functions to ensure that representations do not degenerate into constant vectors. Techniques include whitening the latent space derived from various image augmentations to maintain diversity [22], and optimizing the empirical cross-correlation matrix to mirror the identity matrix in the Barlow Twins framework, thus minimizing input redundancy [21]. Additional strategies involve introducing a loss term that increases the variance among images in a mini-batch beyond a set threshold for stronger outcomes [19], random grouping of feature dimensions for comparison against basis norms [42], and TWIST, which aligns the class distributions of two augmented images through a mutual information maximization constraint in its cost function to avert collapse [43]. The Self-Classifier approach further refines this by adjusting the standard cross-entropy loss with a uniform prior assumption on labels, leading to improved performance [44].

## 3. Proposed Patch-Wise Self-Supervised Representation Learning

**3.1 An Explanation of DINO and Its Limitation**

The recent self-supervised framework DINO, with the aid of knowledge distillation, could utilize the power of Vision Transformers (ViTs) for the self-supervised representation learning problem [18]. In this framework, by performing various kinds of augmentations, the global and local crops of an original image are extracted. Global crops pass through the teacher and both the global and local crops pass through the student network. The idea is to pull representations resulting from the outputs of these two networks. As a result, the CLS embedding, which is the representation of the whole image, is extracted for each augmented image. Then, the cross-entropy loss between the outputs of student and teacher networks is computed to match the representation of the local image with the representation of the global variant.

In this framework, the emphasis is placed solely on the CLS tokens, serving as representatives of the entire input image, while tokens associated with image patches are disregarded. The model captures image-level representations while neglecting patch-level representations. While it is shown that the image-level representation constraint leads the model to learn a meaningful representation, this constraint may not be sufficient for acquiring an exhaustive representation space. For instance, when an image encompasses multiple objects, matching representations of the entire image across augmented views may lead to confusion regarding the primary focus of the image. However, introducing specific constraints, such as ensuring consistency in representations of local regions, could enhance the model's understanding of image components. This approach facilitates a discernible differentiation between object representations. Even if the image contains only a single object, by imposing additional specific restrictions, the model is anticipated to develop an enhanced understanding of the various components of that object. For example, if the image depicts a dog, by pulling representations of the dog's head in the augmented images, the model can distinguish between the head and other parts of the dog more effectively.

**3.2 Proposed Idea: Patch-Wise Local-Global Matching**

Our key idea is to capture local information inside the image patches and use it alongside the global information to train the model more effectively. We obtain local details by identifying and matching corresponding patches in augmented images. This matching process introduces additional loss terms, which we believe act as effective inductive biases. These biases aid the model in achieving a more detailed level of abstraction, a feature that was notably absent in the DINO approach. As a result, our methodology enables the model to grasp not just the broad themes of an image but also to gain a deep insight into its finer details.

Figure 1 depicts the PW-Self concept as follows: An original image undergoes spatial augmentation to create global and local views. These augmentations alter pixel positions through techniques like cropping, resizing, and horizontal flipping. Each view is then cut into patches and experiences individual patch-level photometric augmentations where each patch is augmented by photometric operators independently from others (3.3). The augmented patches are processed by a ViT encoder network, producing CLS and patch-specific tokens. The patch-matching algorithm compares global and local views to identify positive patches (3.4). This process determines matching patches with shared concepts, synchronizing their representations through cross-entropy loss. A loss function consolidates these alignments into a final result (3.5).

Our framework incorporates a dual-network architecture consisting of a student and a teacher network, an approach inspired by the principles of self-distillation [18]. It facilitates effective learning and knowledge transfer within the self-supervised learning paradigm without the need for negative samples. For feeding the images into the student and teacher networks, the global and local crops are resized to $224^2$ and $96^2$ respectively. The global crops are input to the teacher network, while both crop types are fed to the student network for feature extraction. Simultaneous training of the student and teacher networks through backpropagation optimizes the learning process, enhancing the model's ability to understand and represent both the micro and macro aspects of images.

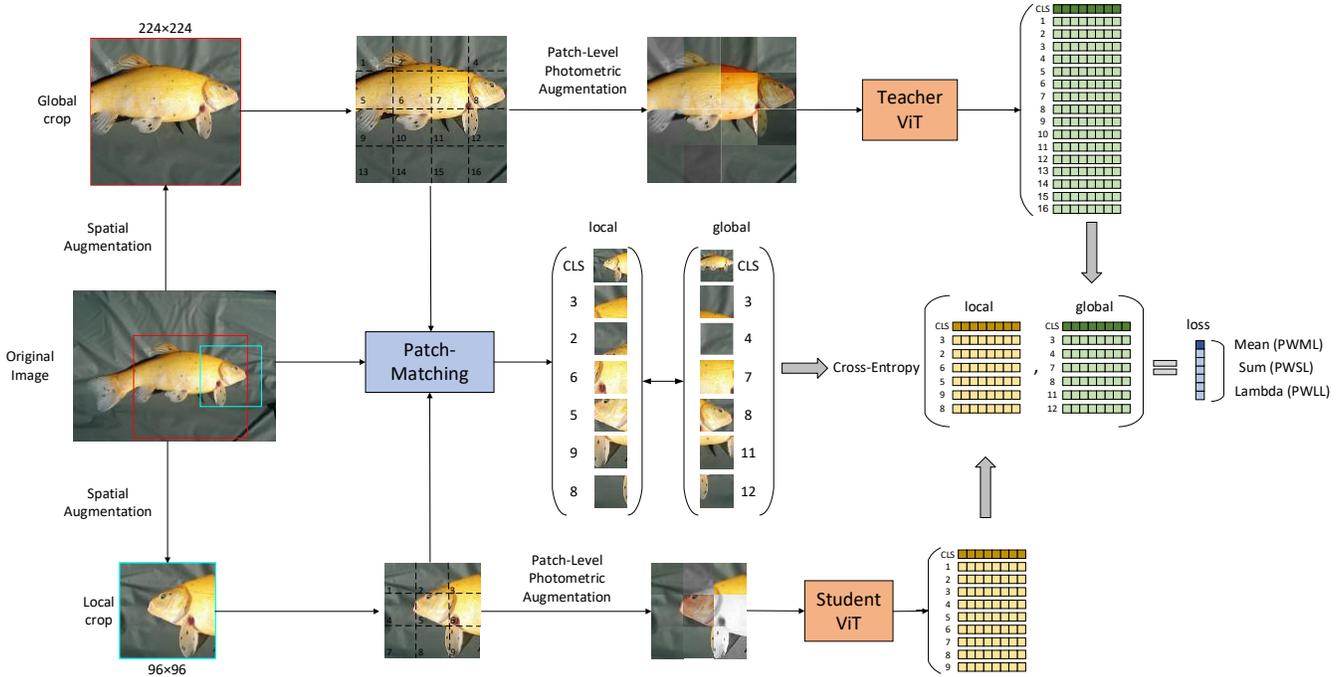

*Figure 1 An original image is spatially augmented via random crop, resize and random flip operations in two ways to obtain global and local crops. Local crop: 0.05<scale<0.4, resize to 96x96, global crop: 0.4<scale<1, resize to 224x224. Then, the two views are cut into patches and undergo patch-level photometric augmentation. Subsequently, they are encoded by two separate Vision Transformers (ViT). Either of the ViT encoders extract representations for the entire image (CLS tokens) as well as each of the patches. PW-Self minimizes the representation distance between CLS tokens as well as corresponding patches across views. Correspondences are guided by the proposed patch-matching algorithm.*

**3.3 Patch-Level Photometric Augmentation**

In traditional approaches, augmentation is applied uniformly across the entire image, leading to a homogenized perspective where local variations are often masked by global transformations. Our method diverges from this norm by treating each patch as an independent entity. It divides views into patches. Each patch is treated as a standalone unit, undergoing its unique augmentation process. This stage is characterized by the exclusive use of photometric augmentations, which alter the appearance of the patches without affecting their spatial position. Photometric augmentations encompass a variety of operators, including color jittering, solarization, blurring, and Gaussian noise. A random selection from these operators is applied to each patch, ensuring that no two patches necessarily experience the same combination of augmentations. This approach allows for the creation of a more complex and varied training dataset, where each patch can exhibit variations that are not necessarily correlated with the global image context. It is particularly advantageous in scenarios involving complex scenes with diverse elements. By exposing the model to a broader spectrum of variations at the patch level, we equip it with a more refined understanding of local features, enhancing its ability to distinguish subtle differences in texture, shape, and color.

**3.4 Patch-Matching Algorithm For Local And Global Crops**

Among the two phases of spatial and photometric augmentations, only the spatial augmentations affect the positions of pixels. As a result, the patch-matching algorithm described in this section only accounts for the operators inside the spatial augmentations which are crop, resize, and horizontal flip. Figure 2 illustrates our patch-matching algorithm applied to an original image subjected to random crop and resize. Identifying corresponding patches between the global and local crops hinges on finding their overlapping area. If there is no overlap, the only matching elements are the CLS tokens. To pinpoint the overlap, we determine which portions of the original image were cropped. By calculating the overlap between the pixel ranges of the local and global crops, we identify the intersection area (depicted as a green rectangle in Figure 2). If $(x_{min}, y_{min}, x_{max}, y_{max})$ and $(x'_{min}, y'_{min}, x'_{max}, y'_{max})$ represents the pixel coordinates of global and local crops respectively, the intersection area is calculated by finding the overlapping pixel ranges between them:

$$intersection\ area = [\max(x_{min}, x'_{min}), \max(y_{min}, y'_{min}), \min(x_{max}, x'_{max}), \min(y_{max}, y'_{max})] \quad (1)$$

The intersection area includes separate patches for the local and global crops. Even if this area does not cover an entire patch at the rectangle's edges, we consider a patch part of the intersection if it includes more than half of its width and height. In the example Figure 2, assuming the patch size 16, the intersection area contains some patches ranging between 21 to 36 from the local and between 1 to 48 from the global crops. These patches are distributed across three rows and four columns in the local and four rows and six columns in the global crops.

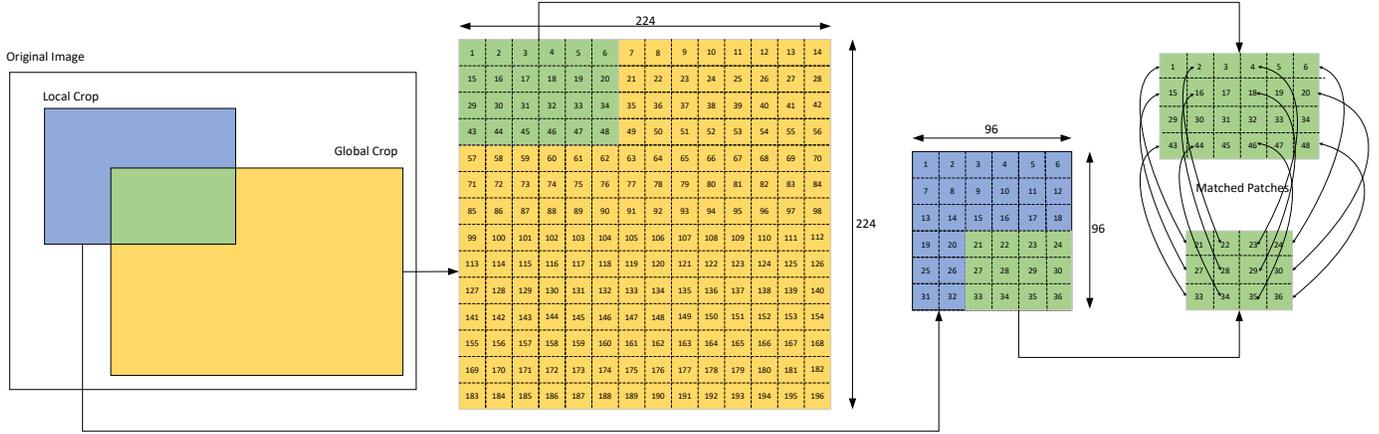

*Figure 2 The original image is augmented in two ways local and global. The local crop (blue rectangle) contains less than 40% of the original image while the global crop (yellow rectangle) contains more than 40% of it. Then, they are resized to 96x96 and 224x224 respectively. To find the corresponding patches across the local and global crops, the intersection area (green rectangle) is determined. The intersection area includes separate patches for the local and global crops. Assuming the patch size is 16, in this example, the intersection area contains 12 patches from the local crop and 24 patches from the global crop. To determine the corresponding patches, we select rows/columns at specific intervals. This interval is calculated by dividing the total number of rows/columns in the global crop by the number of rows/columns in the local crop minus one (Eq ( 2 ), Eq ( 3 )). To select the matching rows/columns, we multiply integer numbers (0, 1, 2, etc.) by the interval and round down to the nearest whole number (Eq( 4 ), Eq( 5 )). As a result, in this example, the 1st, 2nd, and 4th rows, as well as the 1st, 2nd, 4th, and 6th columns of the global crop are selected.*

In the patch-matching algorithm, the aim is to match the patches of one view with the other view uniformly. This is done by selecting some of the patches within the view with more rows/columns to be matched with all patches within the view with fewer rows/columns while ensuring that the first and last rows/columns are matched together. Since the local crop is smaller than the global, it usually contains fewer rows/columns in the intersection area. However, there might be cases where the global crop has fewer rows/columns. To determine the corresponding patches along the rows, we select rows at specific intervals. This interval is calculated by dividing the total number of rows in the global crop by the number of rows in the local crop minus one (Eq( 2 )). The notations used in the following equations are listed in Table 1. The same approach is used for columns as well (Eq( 3 )). However, if the number of rows or columns in the intersection area of the local crop is greater, we calculate the interval by dividing the number of rows or columns in the local crop by the number of rows or columns in the global crop minus one. To select the desired rows or columns, we multiply integer numbers (0, 1, 2, etc.) by the interval and round down to the nearest whole number (Eq( 4 )) and (Eq( 5 )).

*Table 1 Notations description*

| Notations | Description |
|---|---|
| GCR | Number of global crop rows |
| LCR | Number of local crop rows |
| GCC | Number of global crop columns |
| LCC | Number of local crop columns |
| RI | Row interval |
| CI | Column interval |
| NC | Number of columns |
| BS | Batch size |
| MP | Number of matched patches |
| PWML | Patch-Wise Mean Loss |
| PWSL | Patch-Wise Sum Loss |
| PWLL | Patch-Wise Lambda Loss |
| PW-Self | Patch-Wise Self-Supervised Representation Learning |

$$RI = \frac{GCR}{LCR - 1} \qquad (2)$$

$$CI = \frac{GCC}{LCC - 1} \quad (3)$$

$$\text{selected rows} = \lfloor i \times RI \rfloor, i = 0,1,2,\ldots \quad (4)$$

$$\text{selected columns} = \lfloor i \times CI \rfloor, i = 0,1,2,\ldots \quad (5)$$

If a horizontal flip is applied to the local or global crops, it merely reverses the sequence of patches in the affected image. For a patch $P(i,j)$, the flipped position is calculated as:

$$P'(i,j) = P(i, NC - j) \quad (6)$$

### 3.5 Loss Function Setup

Our framework leverages knowledge distillation, an effective technique for training deep learning models. In this setup, we utilize a dual-network architecture comprising a teacher network and a student network, both of which are trained simultaneously. The teacher network is not pretrained; instead, it learns concurrently with the student network. To update the teacher network's parameters efficiently, we employ the Exponential Moving Average (EMA) method. This approach has been explored in various models [8, 18, 30, 45] and is particularly well-suited for our framework, ensuring synchronized and effective learning between the two networks. The update rule is as follows: $\theta_t \leftarrow \lambda \theta_t + (1 - \lambda)\theta_s$ where $\theta_t$, $\theta_s$ denote the parameters of the teacher and student networks respectively, and $\lambda$ follows a cosine schedule. Also, to avoid collapse the centering and sharpening operators have been used. By feeding an image to the network, the probability distribution $P$ results due to the softmax function as the ultimate layer: $P_s(x)^i = \frac{\exp(\theta_s(x)^i/\tau_s)}{\sum_{k=1}^{K} \exp(\theta_s(x)^k/\tau_s)}$ where $i$ denotes the $i^{th}$ dimension of the distribution, $\tau_s$ is the temperature parameter of the student controlling the sharpness of the output distribution. The teacher network shares the same structure with the difference that the centering operator $C$ is also involved: $P_t(x)^i = \frac{\exp((\theta_t(x)^i - C)/\tau_t)}{\sum_{k=1}^{K} \exp((\theta_t(x)^k - C)/\tau_t)}$ where $C$ is the average of samples inside the batch.

If $x_i$ is the $i^{th}$ image in the batch, by performing augmentations on that, $x_i^{g(j)}, x_i^{l(j)}$ are obtained which are the $j^{th}$ global and local crops of $x_i$ respectively. In our implementation, 2 global and 8 local crops are extracted. These 2 global crops are passed through both the teacher and student networks. The results are $g_i^j = P_t(x_i^{g(j)}), j = 1,2$ and $gs_i^j = P_s(x_i^{g(j)}), j = 1,2$ respectively. The 8 local crops are only fed to the student network where the results are $ls_i^j = P_s(x_i^{l(j)}), j = 1,2,\ldots,8$. Hence the outputs of the student network are $l_i^j = [gs_i^1, gs_i^2, ls_i^1, ls_i^2, \ldots, ls_i^8]$; where the first two elements are the results of feeding global crops and the eight remaining are from feeding local crops. Then, the loss between the outputs of teacher and student is calculated using the cross-entropy loss $H(a,b) = -a \log(b)$.

In the DINO loss function (Eq( 7 )), the cross-entropy between the CLS representations (token 0) of the outputs of teacher and student are calculated. Two cases are excluded from the loss calculations ($2 \times 10 - 2 = 18$). Those are when $i = j = 1$ and $i = j = 2$. The reason is that in both cases, the student and teacher have been operated on the same view. The mentioned operation is repeated for every image in the batch. The final loss is the average of the calculated terms. The notations used in the equations are listed in Table 1.

$$\text{DINO } Loss = \frac{\sum_{b=1}^{BS}\sum_{i=1}^{2}\sum_{j=1, j\neq i}^{10} H(g_b^i[0], l_b^j[0])}{BS \times 18} \quad (7)$$

In the Patch-Wise Mean Loss (PWML in Eq( 8 )), the same operations are done with the difference that for every global and local crop, the correspondent patches have been calculated and stored in $g_b^i$ and $l_b^j$. Hence the cross-entropy between all the matched patches (including CLS as well as other tokens) are calculated. The final loss is the average between those terms.

$$PWML = \frac{\sum_{b=1}^{BS}\sum_{i=1}^{2}\sum_{j=1, j\neq i}^{10}(\sum_{p=0}^{MP-1} H(g_b^i[p], l_b^j[p])/MP)}{BS \times 18} \quad (8)$$

In the Patch-Wise Sum Loss (PWSL in Eq( 9 )), the final loss results from the summation of the cross-entropy terms.

$$PWSL = \frac{\sum_{b=1}^{BS}\sum_{i=1}^{2}\sum_{j=1, j\neq i}^{10}\sum_{p=0}^{MP-1} H(g_b^i[p], l_b^j[p])}{BS \times 18} \quad (9)$$

In both PWML (Eq( 8 )) and PWSL (Eq( 9 )), each cross-entropy term contributes to the final loss equally. However, in Patch-Wise $\lambda$ Loss (PWLL in (Eq( 10 )) we differentiate between the CLS tokens and other tokens' weights. This is done by giving weight $\lambda$ to the CLS tokens and weight $1 - \lambda$ to others.

$$PWLL = \frac{\sum_{b=1}^{BS}\sum_{i=1}^{2}\sum_{j=1, j\neq i}^{10}(\lambda \times H(g_b^i[0], l_b^j[0]) + (1 - \lambda) \times \left(\frac{\sum_{p=1}^{MP-1} H(g_b^i[p], l_b^j[p])}{MP - 1}\right))}{BS \times 18} \quad (10)$$

A pseudo-code implementation is proposed in Algorithm 1 for better comprehension of the algorithm.

*Algorithm 1 Patch-Wise Self-Supervised Representation Learning (PW-Self) Pseudocode in Python*

```
Input: Input: A batch of images
Output: Cross-entropy loss of matching CLS tokens as well as correspondent patches

 1: # Synchronize parameters of the teacher (gt) and student (gs) networks:
 2: gt.params := gs.params
 3: for each img in loader do
 4:     # Augment image spatially using crop, resize and flip operators
 5:     # SpatialAugment returns augmented image, cropped coordinates, and flip status
 6:     [img1, coords1, flipped1] := SpatialAugment(img)
 7:     [img2, coords2, flipped2] := SpatialAugment(img)
 8:
 9:     # Find matching patches between two augmented local and global images
10:     correspondences := PatchMatching([img1, coords1, flipped1], [img2, coords2, flipped2])
11:
12:     # Patch-Level Photometric Augmentation using color jittering, solarization, blurring, and Gaussian noise
13:     img1 := PhotometricAugment(img1)
14:     img2 := PhotometricAugment(img2)
15:
16:     # Obtain representations from teacher and student networks:
17:     t := gt(img1)
18:     s := gs(img2)
19:
20:     # Calculate cross-entropy loss for the matching patches based on PWML Eq( 8 )
21:     loss := CalculateLoss(t, s, correspondences, PWML)
22:     loss.backward() # Backpropagate loss
23: end for
24
25: # Calculate loss based on the resulted representations, patch correspondences, and loss function
25: function CalculateLoss(t, s, correspondences, loss_function):
26:     t := t.detach() # Detach teacher outputs to stop gradients
27:     t_patches := t[:, correspondences.crop1, :]
28:     s_patches := s[:, correspondences.crop2, :]
29:
30:     # Compute cross-entropy loss between corresponding patches
31:     CEloss := CrossEntropyLoss(t_patches, s_patches)
32:     loss_tensor := torch.sum(CEloss, dim= -1)
33:
34:     # Aggregate loss using the specified loss (refer to Eq( 8 ), Eq( 9 ), Eq( 10 ))
35:     final_loss := loss_function(loss_tensor)
36:     return final_loss
37: end function
```

## 4. Results

### 4.1 Experimental Setup

**4.1.1 Datasets**

The datasets used in this work are in common with related works in the literature. We have used three datasets Cifar10 [46], ImageNet-100 [47], and ImageNet-1K [47] to pretrain our network. Cifar10 is a small-scale dataset that has 10 classes with images in 32×32 resolutions. It has 60,000 images of which 50,000 are for training and 10,000 for validation. ImageNet-1K is an image dataset that aligns with the WordNet hierarchy. Each synset represents a meaningful concept and is associated with one or more words or word phrases. ImageNet's primary goal is to offer an average of 1,000 curated images for each synset, all of which undergo quality control and human annotation. It encompasses 1000 object classes and comprises 1,281,167 training images, 50,000 validation images, and 100,000 test images. ImageNet-100 is a subset of the ImageNet-1K dataset that has 100 classes randomly selected. The dataset is well-balanced such that each class contains 1,300 training images and 50 validation images.

**4.1.2 Implementation Details**

In order to find the patch correspondences across the global and local crops, it is necessary first to find the intersection area between the global and local crops. For that, it is needed to return the cropped area (start and end pixels for both the width and height), randomly chosen from the original image when performing random crop operation. These parameters are not returned by default in deep learning frameworks *TensorFlow* and *PyTorch*. As a result, we overwrote the original implementation of the method in PyTorch. By doing so, we could find the intersection area between each crop extracted from the original image.

In our proposed method, the patch-matching algorithm identifies corresponding patches for images within a mini-batch, followed by computing cross-entropy loss for each pair of matched patches. This step is typically time-intensive due to its incompatibility with vectorization techniques, unlike the DINO framework, which matches only CLS tokens without seeking matched patches. To address the issue, we implemented a simplification in our algorithm: all images in a batch undergo identical spatial augmentations. This is achieved by enforcing the same random seed for spatial augmentation operators: resize, crop, and horizontal flip across all samples in a batch. For instance, if the second global view of the first image in a batch is cropped at certain points, the same applies to the second global view of all other images in the batch. This uniformity means the patch-matching algorithm needs to be applied only once per batch, with the results applicable to all images. This strategy allows for vectorized loss calculations, leading to a substantial reduction in code execution time while not compromising the overall network's performance.

In this work, the Adamw optimizer [48] with batch sizes from 10 to 40 is used for different datasets. The learning rate undergoes a linear ramp-up phase in the initial 10 epochs until it reaches its base value. It decays based on a cosine schedule. Similarly, weight decay follows a cosine schedule, transitioning from 0.04 to 0.4. The temperature parameter ($\tau_s$) is set at 0.1, and a linear warm-up scheme is applied to the parameter $\tau_t$, transitioning from 0.04 to 0.07 for the initial 30 epochs.

**4.2 Evaluation: Linear and KNN image classification**

Like other self-supervised representation learning methods, the PW-Self method is pretrained on a dataset without considering its labels. Here, we pretrained PW-Self on three datasets Cifar10, ImageNet-100, and ImageNet-1K separately. In the next stage, the images of that dataset are fed to the pretrained network, and the features are extracted to classify the images. The classification is done in two manners: 1) Linear validation: an MLP layer is placed on top of the extracted features. Then, it will be trained to classify the images. 2) KNN validation: The features are extracted and frozen. The K-Nearest Neighbors (KNN) with K=10 or 20 is performed on them.

**4.2.1 Pretraining on Cifar10**

Table 2 illustrates the results of validating the frozen features that come from pretraining the network for classifying the Cifar10 dataset after training for 200 epochs. As it is clear, PW-Self, with any loss function, outperforms DINO and other state-of-the-art methods both in linear validation and KNN validation. An interesting point is that our method (when using PWML (Eq( 8 )) could obtain more increase w.r.t. DINO when doing KNN validation compared to linear validation. By comparing the results of PW-Self with DINO, (rows 3 and 10 in Table 2), the increase obtained in linear validation is +5.86%, while the increase that happened in KNN validation (K=10) is +7.67%. This phenomenon ascertains that in the situation where there is no additional training phase (like KNN validation), the separability of the representations resulting from PW-Self is higher than DINO.

Table 2 The comparison of PW-Self and state-of-the-art methods when pretrained on Cifar10 for 200 epochs. PW-Self is trained with ViT-Base as the backbone. The experiments are done with various configurations for loss function (PWML, PWSL, PWLL), and patch size 16. Linear validation and KNN validation columns demonstrate the Cifar10 image classification accuracies. In all experiments, the PW-Self outperforms existing approaches.

| Pretrain Dataset | Method | Backbone | Patch size | Loss Function | Linear Validation▲ | KNN Validation (K=10) | (K=20) |
|---|---|---|---|---|---|---|---|
| Cifar10 | SCLR [9] | ResNet-50 | | | 86.54 | 82.81 | 82.11 |
| | MoCov2 [29] | ResNet-50 | | | 87.61 | 83.52 | 83.74 |
| | DINO [18] | ViT-Base | 16 | Eq( 7 )) | 87.79 | 84.44 | 84.91 |
| | BarlowT [21] | ResNet-50 | | | 88.74 | 86.14 | 85.26 |
| | OBoW [13] | ResNet-50 | | | 88.81 | 87.64 | 86.17 |
| | SwAV [11] | ResNet-50 | | | 90.13 | 90.41 | 88.84 |
| | BYOL [16] | ResNet-50 | | | 90.41 | 89.32 | 90.87 |
| | EsViT [27] | ViT-Base | 16 | | 90.56 | 90.15 | 88.27 |
| | ARB [42] | ResNet-50 | | | 92.19 | 91.75 | 91.46 |
| | PW-Self | ViT-Base | 16 | PWML (Eq( 8 )) | 93.65 | **92.11** | **92.07** |
| | | | | PWSL (Eq( 9 )) | 92.71 | 91.21 | 90.13 |
| | | | | PWLL ($\lambda = 0.2$, Eq( 10 )) | **93.92** | 91.85 | 91.47 |

**4.2.2 Pretraining on ImageNet-100**

Table 3 presents the results of the comparison of our method and other state-of-the-art approaches when pretrained on ImageNet-100 after training for 200 epochs. The proposed PW-Self method with the PWML loss function (Eq( 8 )) and patch size 16 could outperform other state-of-the-art methods both in linear and KNN validations.

More experiments with various backbone architectures ViT-Tiny, ViT-Small, and ViT-Base are reported in Table 4 to provide more visibility. One interesting point in Table 4 is that the linear validation improvements made by PW-Self compared to DINO have been increased by enlarging the number of training parameters. To explain it more, when ViT-Tiny, ViT-Small, and ViT-Base are used as the backbone, the linear validation improvements of PW-Self w.r.t the DINO are +3.75%, +3.78, and +5.24% each in order. This confirms that the PW-Self, due to the injected inductive biases through the learning process, has more capacity to learn the relations in comparison with the DINO.

*Table 3 The results comparison of PW-Self and state-of-the-art methods when pretrained on ImageNet-100 for 200 epochs. PW-Self is trained with ViT-Base as the backbone, PWML, and patch size 16. Linear validation and KNN validation columns demonstrate the ImageNet-100 image classification accuracies. PW-Self outperforms existing approaches.*

| Pretrain Dataset | Method | Backbone | Patch size | Loss Function | Linear Validation▲ | KNN Validation | |
|---|---|---|---|---|---|---|---|
| | | | | | | (K=10) | (K=20) |
| ImageNet-100 | SCLR [9] | ResNet-50 | | | 71.72 | 55.64 | 55.92 |
| | MoCov2 [29] | ResNet-50 | | | 74.12 | 57.65 | 57.41 |
| | OBoW [13] | ResNet-50 | | | 75.29 | 58.43 | 58.74 |
| | BarlowT [21] | ResNet-50 | | | 76.91 | 58.48 | 58.32 |
| | BYOL [16] | ResNet-50 | | | 77.52 | 59.58 | 59.19 |
| | SwAV [11] | ResNet-50 | | | 77.68 | 61.51 | 61.74 |
| | DINO [18] | ViT-Base | 16 | Eq( 7 ) | 78.22 | 68.58 | 68.84 |
| | ARB [42] | ResNet-50 | | | 79.48 | 73.51 | 73.87 |
| | EsViT [27] | ViT-Base | 16 | | 80.44 | 71.74 | 70.24 |
| | MSBReg [40] | ResNet-18 | | | 81.56 | 74.51 | 75.12 |
| | **PW-Self** | **ViT-Base** | **16** | **PWML (Eq( 8 ))** | **83.46** | **76.37** | **76.23** |

*Table 4 The results comparisons of PW-Self and DINO when pretrained on ImageNet-100 for 200 epochs. Both methods are trained with ViT-Tiny, ViT-Small, and ViT-Base as the backbone, and PWML as the loss function for PW-Self. Linear validation and KNN validation columns demonstrate the ImageNet-100 image classification accuracies. In all experiments, PW-Self outperforms DINO.*

| Pretrain Dataset | Method | Backbone | Patch size | Loss Function | Linear Validation▲ | KNN Validation | |
|---|---|---|---|---|---|---|---|
| | | | | | | (K=10) | (K=20) |
| ImageNet-100 | DINO [18] | ViT-Tiny | 16 | Eq( 7 ) | 58.08 | 51.84 | 52.74 |
| | **PW-Self** | **ViT-Tiny** | **16** | **PWML (Eq( 8 ))** | **61.83** | **59.23** | **59.07** |
| | DINO [18] | ViT-Small | 16 | Eq( 7 ) | 70.92 | 64.64 | 64.74 |
| | **PW-Self** | **ViT-Small** | **16** | **PWML (Eq( 8 ))** | **74.70** | **68.45** | **68.28** |
| | DINO [18] | ViT-Base | 16 | Eq( 7 ) | 78.22 | 68.58 | 68.84 |
| | **PW-Self** | **ViT-Base** | **16** | **PWML (Eq( 8 ))** | **83.46** | **76.37** | **76.23** |

### 4.2.3 Pretraining on ImageNet-1K

We pretrained the proposed PW-Self method with ViT-Base, patch size 16, and the PWML loss function (Eq( 8 )) on ImageNet-1K for 50 epochs. The linear and KNN validation results are reported in Table 5. As can be seen, PW-Self outperforms other existing methods successfully.

*Table 5 The results comparison of PW-Self and state-of-the-art methods when pretrained on ImageNet-1K for 50 epochs. PW-Self is trained with ViT-Base as the backbone, PWML, and patch size 16. Linear validation and KNN validation columns demonstrate the ImageNet-1K image classification accuracies. PW-Self outperforms existing approaches.*

| Pretrain Dataset | Method | Backbone | Patch size | Loss Function | Linear Validation▲ | KNN Validation | |
|---|---|---|---|---|---|---|---|
| | | | | | | (K=10) | (K=20) |
| ImageNet-1K | SCLR [9] | ResNet-50 | | | 60.06 | 52.81 | 51.16 |
| | BarlowT [21] | ResNet-50 | | | 61.02 | 58.71 | 56.59 |
| | ARB [42] | ResNet-50 | | | 62.05 | 63.46 | 62.51 |
| | MoCov2 [29] | ResNet-50 | | | 64.53 | 53.14 | 52.78 |
| | OBoW [13] | ResNet-50 | | | 68.76 | 58.49 | 57.13 |
| | BYOL [16] | ResNet-50 | | | 69.87 | 60.51 | 60.13 |
| | SwAV [11] | ResNet-50 | | | 70.84 | 62.49 | 61.19 |
| | DINO [18] | ViT-Base | 16 | Eq( 7 ) | 71.46 | 64.55 | 64.34 |
| | EsViT [27] | ViT-Base | 16 | | 73.21 | 67.49 | 67.31 |
| | **PW-Self** | **ViT-Base** | **16** | **PWML (Eq( 8 ))** | **73.51** | **68.71** | **68.42** |

## 4.3 Evaluation: Downstream Tasks

The resulting features can be applied in other image applications as well. In this work, two downstream tasks, copy detection, and image retrieval, are selected to assess the generalizability of the representations coming out from the pretrained networks.

### 4.3.1 Copy Detection

This task is to recognize the distortions of types of insertions, blurring, printing, etc. To evaluate the performance of the methods, the mean Average Precision (mAP) is reported on the "strong" subset of the Copydays dataset [49]. Two distinct image sets of 10k and 20k are required to run the task. The 10k set is for distractor images, and the 20k set is for learning the parameters of the whitening operator. When evaluating the networks pretrained on Cifar10 and ImageNet-100, both image sets are randomly sampled from the ImageNet dataset excluding the images contained in the ImageNet-100. Although, when evaluating the network pretrained on ImageNet-1K, the whole ImageNet-1K was used to train those parameters. Table 6 compares the results of the PW-Self with DINO when pretrained on Cifar10, ImageNet-100, and ImageNet-1K, respectively. The PW-Self model surpasses DINO when not using patch-level augmentation. A more detailed discussion about the reasons is available in 5.4.

### 4.3.2 Image Retrieval

The revisited [50] Oxford and Paris [51] datasets were used for the image retrieval task. They both contain 3 splits Easy (E), Medium (M), and Hard (H). We report the mean Average Precision (mAP) on Medium and Hard splits, following DINO's convention [18]. After pretraining with either DINO or PW-Self, the images of the mentioned datasets were fed to the network. Then, the features are extracted and frozen. Finally, the KNN classifier is applied to retrieve the most relevant images. The mAPs for two datasets are reported in Table 6 when pretraining is done on Cifar10, ImageNet-100, and ImageNet-1K, respectively. As can be seen, PW-Self using PWML (Eq( 8 )) outperforms DINO in both Medium and Hard splits in both datasets Oxford and Paris.

*Table 6 The comparison of downstream tasks results between PW-Self and DINO, when pretrained on Cifar10, ImageNet-100, and ImageNet-1K. The downstream tasks are copy detection and image retrieval. The mAP is reported on the "strong" subset of the copydays (copy detection), and Oxford and Paris (image retrieval) datasets.*

| Pretrain Dataset | Method | Backbone | Patch size | Patch-level Aug. | Loss Function | Copy Detection | Image Retrieval | |
|---|---|---|---|---|---|---|---|---|
| | | | | | | | Oxford | Paris |
| Cifar10 | DINO [18] | ViT-Base | 16 | - | Eq( 7 ) | 0.60 | M: 7.52, H: 1.59 | M: 19.44, H: 5.23 |
| | PW-Self | ViT-Base | 16 | ✓ / ✗ | PWML (Eq( 8 )) | 0.54 / 0.64 | M: 8.83, H: 1.73 / M: 8.64, H: 1.65 | M: 21.51, H: 5.64 / M: 21.34, H: 5.28 |
| ImageNet-100 | DINO [18] | ViT-Base | 16 | - | Eq( 7 ) | 0.73 | M: 13.64, H: 2.50 | M: 23.89, H: 7.11 |
| | PW-Self | ViT-Base | 16 | ✓ / ✗ | PWML (Eq( 8 )) | 0.69 / 0.76 | M: 14.87, H: 3.11 / M: 14.12, H: 2.56 | M: 25.41, H: 7.50 / M: 24.60, H: 7.21 |
| ImageNet-1K | DINO [18] | ViT-Base | 16 | - | Eq( 7 ) | 0.79 | M: 25.85, H: 9.13 | M: 54.88, H: 27.84 |
| | PW-Self | ViT-Base | 16 | ✓ / ✗ | PWML (Eq( 8 )) | 0.75 / 0.81 | M: 28.86, H: 9.19 / M: 28.66, H: 9.17 | M: 58.26, H: 29.73 / M: 58.11, H: 29.71 |

## 5. Discussions

In this section, we discuss the computational complexity of our method and compare it with the most similar approaches in the literature. In addition, the effect of the patch size, backbone architectures, proposed loss functions, and lambda parameters in the proposed PW-Self method are explored.

### 5.1 Computational Complexity Analysis

Our methodology extends beyond the standard DINO framework by incorporating three additional modules: patch-level augmentation, patch-matching, and enhanced loss function calculation. The patch-level augmentation module exhibits a computational complexity of $O(T)$, where $T$ represents the total number of patches in views. Our patch-matching algorithm, streamlined for efficiency, operates at a constant time complexity $O(1)$, owing to its reliance solely on the number of rows and columns in the respective views for effective matching. Moreover, the aggregation of local losses into the overall loss is characterized by a complexity of $O(Td)$, with $d$ denoting the dimensionality of the representations. Consequently, the overall complexity of our proposed method is evaluated at $O(Td)$.

In comparison to similar methods such as EsViT [27] and SelfPatch [26], our approach demonstrates greater efficiency. A comparative analysis of computational complexities, as presented in Table 7, reveals that EsViT's approach, which involves identifying the most similar patch to each anchor patch based on maximal cosine similarity, results in a complexity of $O(T^2 d)$. Meanwhile, SelfPatch locates positive pairs by exploring the 8 neighboring patches of the query, leading to a complexity of $O(8Td)$. Additionally, it employs a smaller transformer in its aggregation module to derive final representations that lead to a complexity of $O(T^2 d)$.

*Table 7 The comparison of computational time complexities between EsViT, and SelfPatch which are the most similar methods to the proposed PW-Self method.*

| Method | Computational Complexity |
|---|---|
| EsViT [27] | $O(T^2 d)$ |
| SelfPatch [26] | $O(T^2 d)$ |
| PW-Self | $O(Td)$ |

**5.2 Effect of Backbone Architecture**

The first three rows in Table 10 show results from using PW-Self, pretrained on ImageNet-100, with different backbone architectures, keeping other settings the same. These architectures are ViT-Tiny, ViT-Small, and ViT-Base, with 5.8 million, 22.2 million, and 86 million parameters, respectively. Increasing the number of parameters allows the model to better understand the data, leading to higher classification accuracies with larger backbones. However, having too many parameters can sometimes cause overfitting, where the model memorizes the training data and can not generalize to unseen data. In our study, the significant improvement seen when moving from ViT-Small to ViT-Base (+8.36%) indicates that the model still had space to learn without overfitting, suggesting that the increase in parameters was beneficial.

**5.3 Effect of Patch Size**

The variation in patch size significantly influences model performance, as demonstrated in Table 8 (rows 2 and 5) and Table 10 (rows 3 and 5). Reducing the patch size from 32 to 16 during Cifar10 pretraining improved linear and KNN validation accuracies by +2.37% and +5.34%, respectively. However, further reducing the patch size from 16 to 8 while pretraining on ImageNet-100 decreased accuracies by -4.27% and -5.36%. This indicates that a slight reduction in patch size effectively introduces a form of regularization, improving the model's ability to generalize. However, reducing the patch size too much imposes excessive constraints on the model. This over-constraint hampers its ability to learn and represent the complexity of the data, resulting in underfitting and decreased performance.

**5.4 Effect of Patch-level Augmentation**

In Table 8 (rows 4 and 5), and Table 10 (rows 3 and 4) the beneficial impact of patch-level augmentation on image classification is illustrated. It shows about a 1% increase in linear and KNN validation accuracies for Cifar10 and ImageNet-100. Conversely, Table 9 (rows 1 and 2) reveals varying effects on downstream tasks. While image retrieval benefits from patch-level augmentation, copy detection experiences a significant decline. This decline is likely due to the conflicting goals of the two approaches. Copy detection focuses on identifying image manipulations, such as altered or distorted sections. Patch-level augmentation, however, is designed to make the model less sensitive to changes in patches, particularly those caused by photometric alterations like color shifts or noise addition. This desensitization means the model may overlook the distortions that copy detection seeks to identify. Thus, patch-level augmentation generally serves as a valuable inductive bias in representation learning, except in scenarios like copy detection where such distortion detection is crucial.

**5.5 Effect of the Loss Function**

Table 8 and Table 9 compare the effectiveness of different loss functions pre-trained on Cifar10. The findings suggest that the choice of loss function does not drastically affect the outcomes. Whether for linear and KNN classification or downstream tasks, each loss function can lead to notable performance. The ability to switch loss functions without significantly affecting other performance metrics is particularly valuable from a practical standpoint. Adjusting the loss function according to specific needs can enhance performance in targeted metrics without substantially impairing others. Specifically, the PWML (Eq( 8 )) tends to yield better results in KNN classification and image retrieval tasks, including on the Oxford and Paris datasets. Furthermore, employing the PWLL (Eq( 10 )) with certain λ values (for instance, λ=0.5 or 0.2 as shown in Table 8) results in improved linear validation accuracies compared to those achieved with PWML.

**5.6 Effect of Lambda (λ) value**

Table 8 highlights a critical insight from employing the PW-Self method and its PWLL mechanism (Eq( 10 )): lower λ values significantly boost model performance. When λ is adjusted to 0.8, the CLS token contributes 80% to the overall loss, with other tokens contributing 20%, resulting in a linear validation accuracy of 76.52%. Conversely, reducing λ to 0.2 shifts the contribution to 20% for the CLS token and 80% for other tokens, enhancing linear validation accuracy to 79.65%. This underscores the significance of balancing patch representations with the CLS token's influence, aligning with the paper's core premise that effective learning leverages both the entire image and its constituent patches. The efficacy of this approach is further corroborated by downstream task results in Table 9, where lowering the lambda value to optimal points (0.2 for image retrieval and 0.5 for copy detection) correspondingly elevates the mAP scores.

*Table 8 The results comparison of PW-Self with different configurations when pretrained on Cifar10. PW-Self is trained with ViT-Tiny as the backbone. The experiments are done with various configurations for loss function (PWML, PWSL, PWLL with multiple values for λ), and patch sizes 32 and 16. Linear validation and KNN validation columns demonstrate the Cifar10 image classification accuracies.*

| Pretrain Dataset | Method | Backbone | Patch size | Patch-level Aug. | Loss Function | Linear Validation | KNN Validation (K=10) | (K=20) |
|---|---|---|---|---|---|---|---|---|
| Cifar10 | DINO [18] | ViT-Tiny | 32 | ✘ | Eq( 7 ) | 71.05 | 72.19 | 72.67 |
| | PW-Self | ViT-Tiny | 32 | ✘ | PWML (Eq( 8 )) | 75.73 | 77.17 | 77.24 |
| | DINO [18] | ViT-Tiny | 16 | ✘ | Eq( 7 ) | 75.29 | 76.38 | 76.24 |
| | PW-Self | ViT-Tiny | 16 | ✓ | PWML (Eq( 8 )) | 79.23 | 83.68 | 83.56 |
| | | | | ✘ | PWML (Eq( 8 )) | 78.10 | 82.51 | 82.41 |
| | | | | ✘ | PWSL (Eq( 9 )) | 78.58 | 78.86 | 78.52 |
| | | | | ✘ | PWLL ($\lambda = 0.9$, Eq( 10 )) | 75.87 | 76.82 | 76.64 |
| | | | | ✘ | PWLL ($\lambda = 0.8$, Eq( 10 )) | 76.52 | 76.65 | 76.94 |
| | | | | ✘ | PWLL ($\lambda = 0.5$, Eq( 10 )) | 79.28 | 79.10 | 79.12 |
| | | | | ✘ | PWLL ($\lambda = 0.2$, Eq( 10 )) | 79.65 | 79.48 | 79.42 |
| | | | | ✘ | PWLL ($\lambda = 0.1$, Eq( 10 )) | 79.06 | 79.36 | 79.04 |

*Table 9 The downstream tasks result comparison of PW-Self with different configurations when pretrained on Cifar10. PW-Self is trained with ViT-Tiny as the backbone. The experiments are done with various configurations for loss function (PWML, PWSL, PWLL with multiple values for λ), and patch size 16. The mAP is reported for copy detection and image retrieval downstream tasks.*

| Pretrain Dataset | Method | Backbone | Patch size | Patch-level Aug. | Loss Function | Copy Detection | Image Retrieval Oxford | Paris |
|---|---|---|---|---|---|---|---|---|
| Cifar10 | PW-Self | ViT-Tiny | 16 | ✓ | PWML (Eq( 8 )) | 0.50 | M: 8.41, H: 1.72 | M: 22.04, H: 5.13 |
| | | | | ✘ | PWML (Eq( 8 )) | 0.59 | M: 8.34, H: 1.63 | M: 21.18, H: 5.11 |
| | | | | ✘ | PWSL (Eq( 9 )) | 0.53 | M: 7.37, H: 1.47 | M: 19.28, H: 4.26 |
| | | | | ✘ | PWLL ($\lambda = 0.9$, Eq( 10 )) | 0.58 | M: 6.35, H: 1.61 | M: 17.71, H: 4.57 |
| | | | | ✘ | PWLL ($\lambda = 0.8$, Eq( 10 )) | 0.59 | M: 6.58, H: 1.52 | M: 17.85, H: 4.57 |
| | | | | ✘ | PWLL ($\lambda = 0.5$, Eq( 10 )) | 0.59 | M: 6.72, H: 1.44 | M: 17.51, H: 4.41 |
| | | | | ✘ | PWLL ($\lambda = 0.2$, Eq( 10 )) | 0.58 | M: 7.12, H: 1.63 | M: 19.37, H: 4.68 |
| | | | | ✘ | PWLL ($\lambda = 0.1$, Eq( 10 )) | 0.57 | M: 6.90, H: 1.47 | M: 20.15, H: 4.84 |

*Table 10 The results comparison of PW-Self with different configurations when pretrained on ImageNet-100. PW-Self is trained with ViT-Tiny, ViT-Small, and ViT-Base as the backbone. The experiments are done with PWML, and patch sizes 16 and 8. Linear validation and KNN validation columns demonstrate the ImageNet-100 image classification accuracies.*

| Pretrain Dataset | Method | Backbone | Patch size | Patch-level Aug. | Loss Function | Linear Validation | KNN Validation (K=10) | (K=20) |
|---|---|---|---|---|---|---|---|---|
| ImageNet-100 | PW-Self | ViT-Tiny | 16 | ✘ | PWML (Eq( 8 )) | 61.52 | 58.66 | 58.10 |
| | | ViT-Small | 16 | ✘ | | 74.44 | 67.92 | 68.16 |
| | | ViT-Base | 16 | ✘ | | 82.80 | 75.24 | 75.43 |
| | | ViT-Base | 16 | ✓ | | 83.46 | 76.37 | 76.23 |
| | | ViT-Base | 8 | ✘ | | 78.53 | 69.88 | 68.78 |

## 6. Conclusion

In this paper, we presented PW-Self, a patch-wise representation learning algorithm, which explored the benefits of considering patches within the self-supervised representation learning frameworks. Our approach uniquely focuses on minimizing the representation distance between corresponding local regions, or patches, thereby adopting a fine-grained perspective that enriches model insights into the intricate details within images. This idea served as an inductive bias that could provide the model with useful insights into the details within the images. To generate more diversity in the training images, the augmented crops were passed through a patch-level augmentation module, wherein individual patches were augmented independently, ensuring distinct and varied transformations separate from those applied to other patches. The alignment of corresponding patches among augmented views was achieved through the application of a simple, yet efficient, patch-matching algorithm. Patch-matching algorithm, due to its efficient design, lowers the computational time complexity compared to similar models. As a result, the model not only gains a profound understanding of the details within images but also operates more quickly compared to traditional patch-wise representation learning methods. Our exhaustive experimental assessments across varied datasets validate the effectiveness of our proposed approach. The patch-wise representation learning algorithm showcases its potential by elevating the state-of-the-art in self-supervised representation learning. This is evidenced by improvements in image classification accuracy: an increase of 1.73% on Cifar10, 1.9% on ImageNet-100, and 0.3% on ImageNet-1K. Furthermore, the method demonstrated promising results in downstream tasks such as copy detection and image retrieval.

## Declarations

### Conflict of interests

The authors declare that they have no conflict of interest.

### Funding

This research did not receive any specific grant from funding agencies in the public, commercial, or not-for-profit sectors.

### Data Availability Statement

The data that support the findings of this study are ImageNet-1K [47], ImageNet-100 [47], and Cifar10 [46], Copydays [49], revisited [50] Oxford and Paris datasets [51] which are openly available.

### Declaration of Generative AI and AI-assisted technologies in the writing process

During the preparation of this work, the authors used ChatGPT to improve the quality of the manuscript writing. After using this tool, the authors reviewed and edited the content as needed. They take full responsibility for the content of the publication.

## REFERENCES

[1] Wu, M., et al. *Conditional negative sampling for contrastive learning of visual representations*. in *International Conference on Learning Representations (ICLR)*. 2021.

[2] Robinson, J., et al. *Contrastive learning with hard negative samples*. in *International Conference on Learning Representations (ICLR)*. 2021.

[3] Kotar, K., et al. *Contrasting contrastive self-supervised representation learning pipelines*. in *Proceedings of the IEEE/CVF International Conference on Computer Vision*. 2021.

[4] Dwibedi, D., et al. *With a little help from my friends: Nearest-neighbor contrastive learning of visual representations*. in *Proceedings of the IEEE/CVF International Conference on Computer Vision*. 2021.

[5] Xiong, Y., M. Ren, and R. Urtasun. *Loco: Local contrastive representation learning*. in *Advances in neural information processing systems (NeurIPS)*. 2020.

[6] Tian, Y., D. Krishnan, and P. Isola. *Contrastive multiview coding*. in *European conference on computer vision (ECCV)*. 2020. Springer.

[7] Kalantidis, Y., et al. *Hard negative mixing for contrastive learning*. in *Advances in Neural Information Processing Systems (NeurIPS)*. 2020.

[8] He, K., et al. *Momentum Contrast for Unsupervised Visual Representation Learning*. in *IEEE/CVF Conference on Computer Vision and Pattern Recognition (CVPR)*. 2020.

[9] Chen, T., et al. *A simple framework for contrastive learning of visual representations*. in *International Conference on Machine Learning (ICML)*. 2020. PMLR.

[10] Oord, A.v.d., Y. Li, and O. Vinyals. *Representation learning with contrastive predictive coding*. in *Advances in Neural Information Processing Systems (NeurIPS)*. 2018.

[11] Caron, M., et al. *Unsupervised learning of visual features by contrasting cluster assignments*. in *Advances in Neural Information Processing Systems (NeurIPS)*. 2020.

[12] Li, J., et al. *Prototypical Contrastive Learning of Unsupervised Representations*. in *International Conference on Learning Representations (ICLR)*. 2020.

[13] Gidaris, S., et al. *Obow: Online bag-of-visual-words generation for self-supervised learning*. in *Proceedings of the IEEE/CVF Conference on Computer Vision and Pattern Recognition (CVPR)*. 2021.

[14] Yan, X., et al. *Clusterfit: Improving generalization of visual representations*. in *Proceedings of the IEEE/CVF Conference on Computer Vision and Pattern Recognition (CVPR)*. 2020.

[15] Caron, M., et al. *Deep clustering for unsupervised learning of visual features*. in *Proceedings of the European Conference on Computer Vision (ECCV)*. 2018.

[16] Grill, J.-B., et al. *Bootstrap your own latent: A new approach to self-supervised learning*. in *Advances in Neural Information Processing Systems (NeurIPS)*. 2020.

[17] Chen, X. and K. He. *Exploring simple siamese representation learning*. in *Proceedings of the IEEE/CVF Conference on Computer Vision and Pattern Recognition (CVPR)*. 2021.

[18] Caron, M., et al. *Emerging Properties in Self-Supervised Vision Transformers*. in *Proceedings of the IEEE International Conference on Computer Vision (ICCV)*. 2021.

[19] Bardes, A., J. Ponce, and Y. LeCun. *Vicreg: Variance-invariance-covariance regularization for self-supervised learning*. in *International Conference on Learning Representations (ICLR)*. 2022.

[20] Zhang, S., et al. *Zero-CL: Instance and Feature decorrelation for negative-free symmetric contrastive learning*. in *International Conference on Learning Representations (ICLR)*. 2021.

[21] Zbontar, J., et al. *Barlow twins: Self-supervised learning via redundancy reduction*. in *International Conference on Machine Learning (ICML)*. 2021. PMLR.

[22] Ermolov, A., et al. *Whitening for self-supervised representation learning*. in *International Conference on Machine Learning (ICML)*. 2021. PMLR.

[23] Azabou, M., et al., *Mine your own view: Self-supervised learning through across-sample prediction*. CoRR, 2021. **abs/2102.10106**.

[24] Bachman, P., R.D. Hjelm, and W. Buchwalter. *Learning representations by maximizing mutual information across views*. in *Advances in Neural Information Processing Systems (NeurIPS)*. 2019.

[25] Zhang, T., et al. *Leverage your local and global representations: A new self-supervised learning strategy*. in *Proceedings of the IEEE/CVF Conference on Computer Vision and Pattern Recognition*. 2022.

[26] Yun, S., et al. *Patch-Level Representation Learning for Self-Supervised Vision Transformers*. in *Proceedings of the IEEE/CVF Conference on Computer Vision and Pattern Recognition (CVPR)*. 2022.

[27] Li, C., et al. *Efficient self-supervised vision transformers for representation learning*. in *Proceedings of the International Conference on Learning Representations (ICLR)*. 2022.

[28] Chen, T., et al. *Big self-supervised models are strong semi-supervised learners*. in *Advances in Neural Information Processing Systems (NeurIPS)*. 2020.

[29] Chen, X., et al., *Improved baselines with momentum contrastive learning*. CoRR, 2020. **abs/2003.04297**.

[30] Chen, X., S. Xie, and K. He. *An empirical study of training self-supervised vision transformers*. in *Proceedings of the IEEE/CVF Conference on Computer Vision and Pattern Recognition (CVPR)*. 2021.

[31] Jang, J., et al. *Self-distilled self-supervised representation learning*. in *Proceedings of the IEEE/CVF Winter Conference on Applications of Computer Vision*. 2023.

[32] Tian, Y., et al. *What makes for good views for contrastive learning?* in *Advances in Neural Information Processing Systems (NeurIPS)*. 2020.

[33] Peng, X., et al. *Crafting better contrastive views for siamese representation learning*. in *Proceedings of the IEEE/CVF Conference on Computer Vision and Pattern Recognition (CVPR)*. 2022.

[34] Wang, W., et al. *Instance-wise Hard Negative Example Generation for Contrastive Learning in Unpaired Image-to-Image Translation*. in *Proceedings of the IEEE/CVF Conference on Computer Vision and Pattern Recognition (CVPR)*. 2021.

[35] Huynh, T., et al. *Boosting contrastive self-supervised learning with false negative cancellation*. in *Proceedings of the IEEE/CVF winter conference on applications of computer vision*. 2022.

[36] Long, X., H. Du, and Y. Li, *Two momentum contrast in triplet for unsupervised visual representation learning*. Multimedia Tools and Applications, 2023: p. 1-14.

[37] Xie, J., R. Girshick, and A. Farhadi. *Unsupervised deep embedding for clustering analysis*. in *International Conference on Machine Learning (ICML)*. 2016. PMLR.

[38] Mo, S., Z. Sun, and C. Li. *Multi-level contrastive learning for self-supervised vision transformers*. in *Proceedings of the IEEE/CVF Winter Conference on Applications of Computer Vision*. 2023.

[39] Hsieh, C.-Y., et al. *Self-Supervised Pyramid Representation Learning for Multi-Label Visual Analysis and Beyond*. in *Proceedings of the IEEE/CVF Winter Conference on Applications of Computer Vision*. 2023.

[40] Moon, S., et al. *An Embedding-Dynamic Approach to Self-Supervised Learning*. in *Proceedings of the IEEE/CVF Winter Conference on Applications of Computer Vision*. 2023.

[41] Jing, L., et al., *Understanding dimensional collapse in contrastive self-supervised learning*. arXiv preprint arXiv:2110.09348, 2021.

[42] Zhang, S., et al. *Align Representations with Base: A New Approach to Self-Supervised Learning*. in *Proceedings of the IEEE/CVF Conference on Computer Vision and Pattern Recognition (CVPR)*. 2022.

[43] Wang, F., et al., *Self-Supervised Learning by Estimating Twin Class Distributions*. CoRR, 2021. **abs/2110.07402**.

[44] Amrani, E. and A. Bronstein. *Self-Supervised Classification Network*. in *European Conference on Computer Vision (ECCV)*. 2022.

[45] Oquab, M., et al., *Dinov2: Learning robust visual features without supervision*. arXiv preprint arXiv:2304.07193, 2023.

[46] Krizhevsky, A. and G. Hinton, *Learning multiple layers of features from tiny images*. 2009.

[47] Russakovsky, O., et al., *Imagenet large scale visual recognition challenge*. International journal of computer vision, 2015. **115**: p. 211-252.

[48] Loshchilov, I. and F. Hutter, *Fixing weight decay regularization in adam*. 2018.

[49] Douze, M., et al. *Evaluation of gist descriptors for web-scale image search*. in *Proceedings of the ACM international conference on image and video retrieval*. 2009.

[50] Radenović, F., et al. *Revisiting oxford and paris: Large-scale image retrieval benchmarking*. in *Proceedings of the IEEE conference on computer vision and pattern recognition*. 2018.

[51] Philbin, J., et al. *Lost in quantization: Improving particular object retrieval in large scale image databases*. in *2008 IEEE conference on computer vision and pattern recognition*. 2008. IEEE.